\pdfoutput=1

\documentclass[11pt]{article}

\usepackage{EMNLP2023}

\usepackage{times}
\usepackage{latexsym}

\usepackage[T1]{fontenc}

\usepackage[utf8]{inputenc}

\usepackage{microtype}

\usepackage{inconsolata}

\newcommand{\muse}{MuSE}
\newcommand{\ct}{CrossTransformer}
\usepackage{graphicx}
\usepackage{float}
\usepackage{booktabs}
\usepackage{tabularx}
\usepackage{multirow}
\usepackage{amsmath}
\usepackage{bm}
\usepackage{arydshln}
\usepackage{amsfonts}
\usepackage{pgfplots}
\pgfplotsset{compat=1.7}
\usepackage{subcaption}
\usepackage{pifont}
\usepackage{colortbl}
\usepackage{algorithmic}
\usepackage[ruled,vlined]{algorithm2e}

\title{Exchanging-based Multimodal Fusion with Transformer}

\author{Renyu Zhu\textsuperscript{1\thanks{\ \ Equal contribution}}\ \ \ \ Chengcheng Han\textsuperscript{2\footnotemark[1]}\ \ \ \ Yong Qian\textsuperscript{3}\ \ \ \ Qiushi Sun\textsuperscript{4} \\
\bf{Xiang Li\textsuperscript{2\thanks{\ \ Corresponding author}}}\ \ \ \ \bf{Ming Gao\textsuperscript{2}} \ \ \ \ \bf{Xuezhi Cao\textsuperscript{3}}\ \ \ \ \bf{Yunsen Xian\textsuperscript{3}} \\
\textsuperscript{1}NetEase~Fuxi~AI~Lab, Hangzhou, Zhejiang, China\\
    \textsuperscript{2}School~of~Data~Science~and~Engineering, East~China~Normal~University, Shanghai, China \\  \textsuperscript{3}Meituan~Inc., Beijing, China \quad \textsuperscript{4}National~University~of~Singapore \\
    \texttt{zhurenyu@corp.netease.com} \quad \texttt{52215903007@stu.ecnu.edu.cn}  
    \\\texttt{sarahyongq@gmail.com} \quad \texttt{qiushisun@u.nus.edu}
    \\  \texttt{\{xiangli,mgao\}@dase.ecnu.edu.cn} \quad \texttt{\{caoxuezhi, yunsenxian\}@meituan.com}
    \\}

\begin{document}
\maketitle

\begin{abstract}
We study the problem of multimodal fusion in this paper.
Recent 
exchanging-based methods have been proposed 
for vision-vision fusion,
which aim to exchange embeddings learned from one modality to the other.
However,
most of them
project inputs of multimodalities into different low-dimensional spaces and cannot be applied to the sequential input data.
To solve these issues,
in this paper,
we propose a novel exchanging-based multimodal fusion model \muse\ for text-vision fusion based on Transformer.
We first use two encoders to
separately map multimodal inputs into different low-dimensional spaces.
Then we 
employ two decoders to regularize the embeddings and pull them into the same space.
The two decoders 
capture the correlations between texts and images with 
the image captioning task and the text-to-image generation task, respectively.
Further,
based on the regularized embeddings,
we present \ct,
which 
uses two Transformer encoders with shared parameters as the backbone model
to exchange knowledge between multimodalities.
Specifically,
\ct\ first 
learns the global contextual information of the inputs in the shallow layers. 
After that,
it performs inter-modal exchange 
by
selecting a proportion of tokens in one modality
and 
replacing their embeddings with the average of embeddings in the other modality. 
We conduct extensive experiments to evaluate the performance of \muse\ on
the Multimodal Named Entity Recognition task and the Multimodal Sentiment Analysis task.
Our results show the superiority of \muse\ against other competitors.
Our code and data are provided at \url{https://github.com/RecklessRonan/MuSE}.

\end{abstract}

\section{Introduction}
Humans perceive the world by simultaneously processing 
various modalities, 
such as visions, sounds and feelings,
and further fusing them.
Therefore,
\textit{multimodal fusion} has become a key research problem~\cite{baltruvsaitis2018multimodal} in the area of artificial intelligence. 
With the recent success of deep learning,  
a series of 
deep multimodal fusion methods~\cite{bayoudh2021survey, guo2019deep} 
have been proposed and 
applied
on various multimodal tasks, 
such as Multimodal Named Entity Recognition (MNER)~\cite{sun2021rpbert, wang2020deep}
and Multimodal Sentiment Analysis (MSA)~\cite{colombo2021improving, wu2021text}.
Early 
methods mainly can be divided into two categories: aggregation-based and alignment-based. Aggregation-based methods first 
represent each modality by a sub-network and then aggregate different representations with various operators, 
such as concatenation~\cite{zeng2019deep}, averaging~\cite{hazirbas2016fusenet}, and self-attention~\cite{valada2020self}.
Further,
alignment-based methods~\cite{colombo2021improving, song2020modality} adopt a regularization loss to align the embeddings of different sub-networks without explicit combination. 

\begin{figure*}[h]
    \centering
    \includegraphics[width=0.99\textwidth]{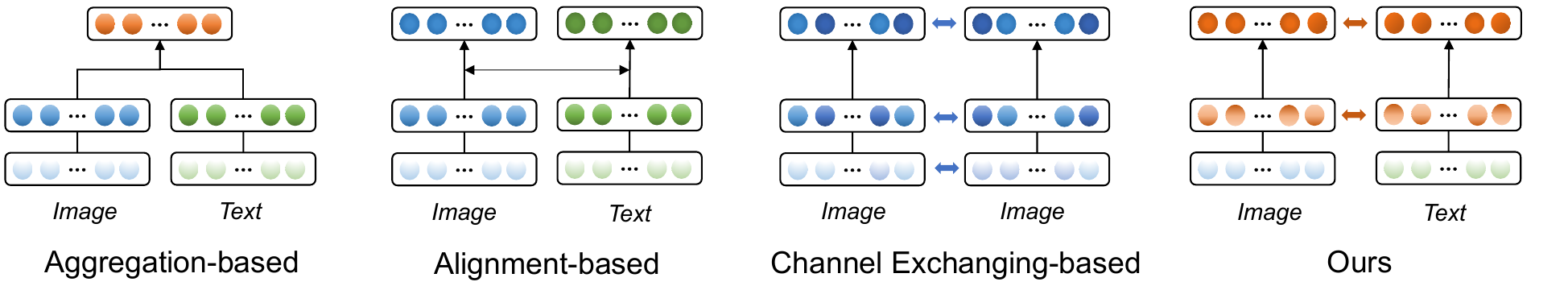}
    \caption{A sketched comparison between existing fusion methods and ours.}
    \label{fig:intro}
\end{figure*}

Recently,
a novel exchanging-based method
CEN~\cite{wang2020deep}
was proposed to handle the trade-off between intra-modal processing and inter-modal fusion. 
By utilizing the scaling factor of Batch Normalization (BN)~\cite{ioffe2015batch}
as the importance measurement of each channel, 
CEN replaces those channels with 
close-to-zero
factor values in one modality with the average of channels in the other modality.
However, 
the method is specially designed for vision-vision fusion and 
the channel exchange
cannot be directly applied 
in other 
multimodal scenarios, such as text-vision fusion.
There are two main challenges.
On the one hand,
CEN implicitly assumes that the two modalities are represented in the same low-dimensional embedding space,
while 
the modalities of texts and images are distant from each other and generally correspond to different spaces.
On the other hand,
the exchanging mode based on CNN channels of images is inapplicable for texts, 
since texts are sequences of words.
Therefore, 
a {research} question arises: \textit{Can we develop an effective exchanging-based neural network model that fuses modalities of text and 
vision?}


In this paper, 
we present a novel \textbf{Mu}ltimodal fu\textbf{S}ion method based on \textbf{E}xchanging, namely, \muse, 
which bridges the gap of exchanging-based methods in the field of text-vision multimodal fusion.
We summarize the comparison between existing multimodal fusion methods and ours in Figure~\ref{fig:intro}.
To handle the problem of different low-dimensional
spaces 
corresponding to various data modalities,
we 
first perform separate low-dimensional projections for texts and images, 
and then
propose two tasks to pull their embeddings into the same space.
Specifically,
inspired by the fact that
texts can be captions of images and images can also be generated from  texts,
we specially design an \emph{image captioning} task and a \emph{text-to-image generation} task
to capture the correlations between texts and images.
We implement the process with an architecture of two 
encoder-decoders,
where encoders are used for text (image) projection and decoders are for image (text) generation.
The two decoders jointly regularize the embeddings generated by the encoders and pull them into the same space.
Further,
we investigate information exchange for text-vision modalities and 
propose \emph{\ct},
which uses two Transformer~\cite{vaswani2017attention} encoders with shared parameters as the backbone model. 
Specifically,
we insert \textit{cls}\footnote{The special token in Transformer that can be used to derive the sentence-level embedding.} into the beginning of the generated embedding vectors of texts and images by the encoder-decoder module,
which are then fed into the two Transformer encoders 
for both modalities.
Intuitively,
we first need to learn 
the global contextual information of the input vectors in 
the shallow layers of the Transformer encoders 
and then exchange the well-learned knowledge between modalities.
After the multimodal fusion completes,
the exchange can stop.
Therefore, 
in \ct,
we introduce two hyper-parameters $\mu$ and $\eta$,
which denote the start layer and the end layer for information exchanging, respectively.
In each exchanging layer,
we 
select the tokens in one modality with the smallest attention scores to \textit{cls}
and replace their embeddings with the average of all token embeddings 
in the other modality.
In this way, 
text-vision fusion can be automatically performed.
Finally, we summarize our contributions as follows.



\begin{itemize}
    \item We 
    generalize the exchanging-based methods from vision-vision fusion to text-vision fusion and
    propose a novel exchanging-based model \muse.
    \item We 
    employ the image captioning task and the text-to-image generation task to capture the correlations between texts and images, which jointly regularize multimodal embeddings and pull them into the same space.
    We further design \ct, which exchanges knowledge between texts and images.
    \item We conduct extensive experiments to compare \muse\ with other state-of-the-arts on the MNER and MSA tasks. Experimental results show the effectivenss of our method.
\end{itemize}

\section{Related Works}
\subsection{Deep Multimodal Fusion}
\label{related_works_fusion}
Deep multimodal fusion has recently received significant attention in deep multimodal learning~\cite{bayoudh2021survey, guo2019deep}. 
Early 
methods can be mainly divided into two categories~\cite{baltruvsaitis2018multimodal}, namely aggregation-based methods~\cite{hazirbas2016fusenet, zeng2019deep, valada2020self} and alignment-based methods~\cite{colombo2021improving, song2020modality}.
For
the former, 
they learn separate representations in each modality and directly aggregate the learned representations from different modalities, 
which lacks the inter-modal communication. 
In contrast, 
alignment-based methods use regularization to align embedding distributions in different modalities, such as mutual information maximization~\cite{colombo2021improving} and maximum mean discrepancy~\cite{gretton2012kernel, shankar2021neural}. However, 
these methods omit the intra-modal characteristics with simple distribution aligning~\cite{song2020modality}. 
While
some works~\cite{han2021improving, ju2021joint} combine aggregation and alignment,
they need {fine-grained} hierarchical design that could
introduce the additional cost of computation and engineering. 
Another taxonomy of multimodal fusion~\cite{de2017modulating} concentrates on the fusion stage and 
a detailed analysis 
is given in~\cite{nagrani2021attention}.
Different from all these methods, 
an exchanging-based method CEN~\cite{wang2020deep} is proposed for multi-vision modalities, which can
effectively handle the trade-off between inter-modal processing and intra-modal fusion. 
While there exist some extensions~\cite{wang2021channel, jiang2022self} to CEN,
all these methods are limited in channel-exchanging and vision-vision fusion. 

Due to the space limitation, we move the related works of MNER and MSA to the Appendix~\ref{related_works_appendix}.

\begin{figure*}[h]
    \centering
    \includegraphics[width=0.98\textwidth]{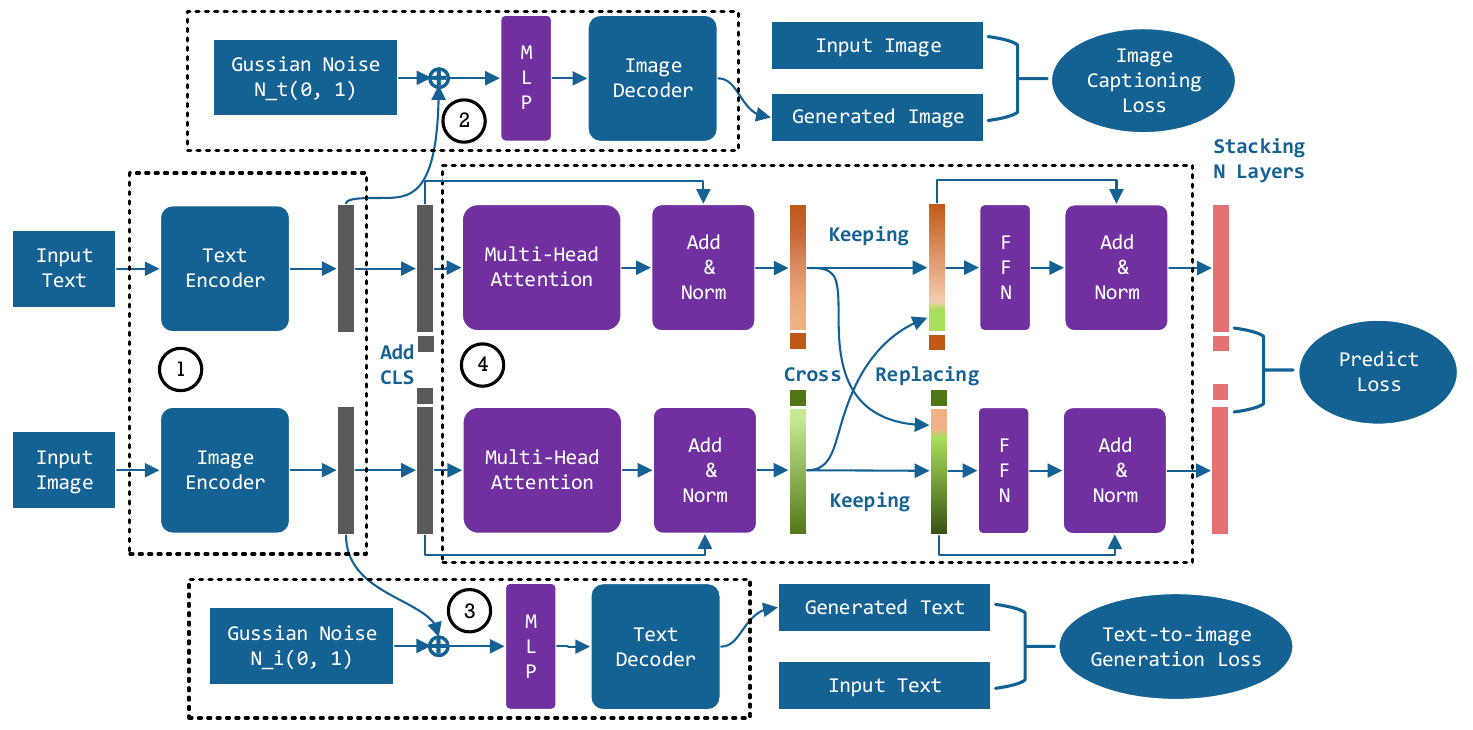}
    \caption{The overall architecture of \muse.}
    \label{fig:main}
\end{figure*}


\section{Methodology}

In this section,
we propose a deep multimodal fusion model 
\muse. 
As shown in Figure~\ref{fig:main}, 
\muse\ is mainly composed of four functional components and we tag them by~\ding{172}-\ding{175}. 
The component~\ding{172} is used to project the input texts and images into low-dimensional spaces,
which includes a text encoder and an image encoder.
Considering that multimodal data might be mapped into different spaces,
we 
further present two embedding regularizers (see components~\ding{173} and~\ding{174}) to pull embeddings of multimodal inputs into the same space. 
These two regularizers are taken as decoders,
which implement the text-to-image generation task and the image captioning task, respectively.
After the embeddings of multimodal inputs are generated,
we feed them into component~\ding{175},
which is a Transformer-encoder-based module called \ct. 
\ct\ performs multimodal information exchanging and finally generates fused embeddings from multimodality. 
Details of each component will be introduced in the following.

\subsection{
Low-dimensional Projection and Embedding Regularization}
We use an architecture of two encoder-decoders to project the inputs of textual ($T$) and visual ($I$) modalities into the same embedding space before exchanging-based fusion.

\paragraph{Low-dimensional Projection.}
We first employ two separate encoders to encode the input texts and images into low-dimensional {embeddings}, respectively:
\begin{align}
\mathbf{T_e} &= \text{TextEncoder}(T),\\
\mathbf{I_e} & = \text{ImageEncoder}(I), \label{encoder}
\end{align}
where TextEncoder can be a typical text representation model (e.g., BERT~\cite{kenton2019bert}) and ImageEncoder can be a typical CNN model (e.g., ResNet~\cite{he2016deep}).
Specifically,
we set $\mathbf{T_e},\mathbf{I_e} \in \mathbb{R}^{n\times d}$, where $n$ is the text length and $d$ is the embedding dimensionality.

\begin{figure}[h]
    \centering
    \includegraphics[width=0.48\textwidth]{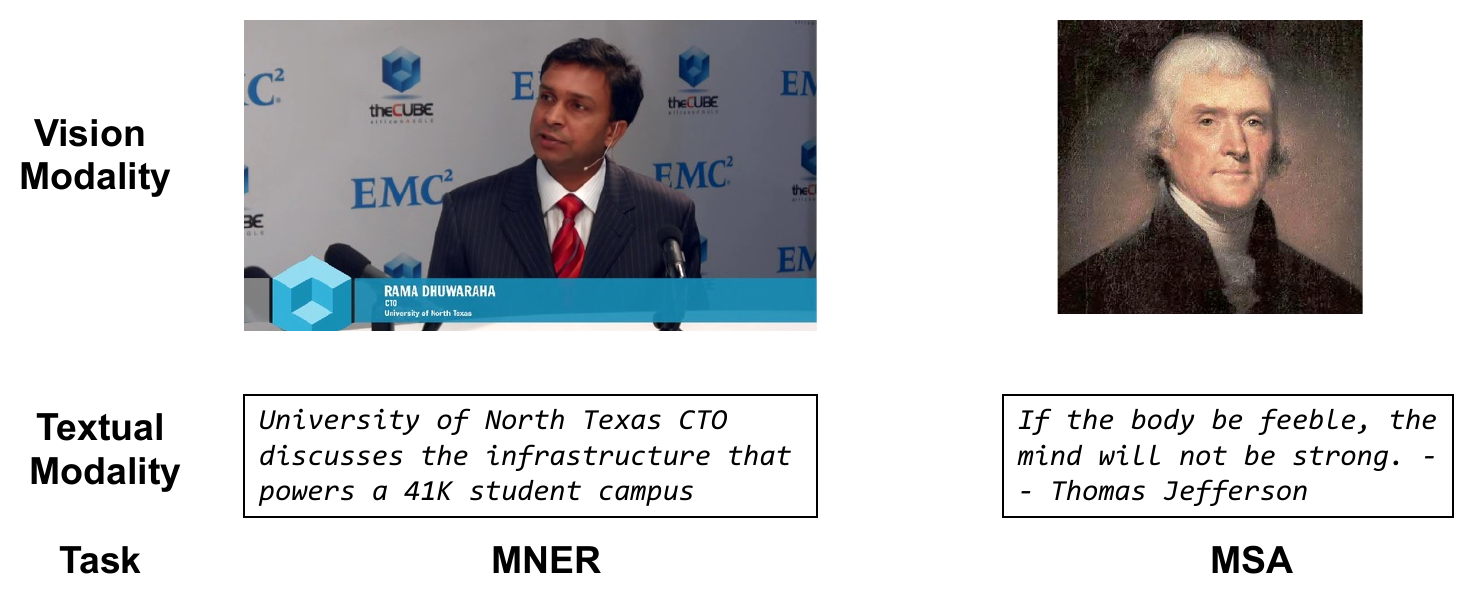}
    \caption{Toy examples on the correlations between texts and images,
    which are extracted from the  
    Twitter15 (for the MNER task) and MVSA-Single (for the MSA task) datasets,
    respectively.}
    \label{fig:example_pair}
\end{figure}

\paragraph{Embedding Regularization.}
Since the inputs are in two modalities,
they are generally projected into different spaces by Equation~\ref{encoder}.
Therefore, 
for information exchange from multimodalities, 
we first need to pull these embeddings into the same space.
Empirically,
we observe that 
texts can be captions of images while images can be generated from texts, as shown in Figure~\ref{fig:example_pair}.
To capture the correlations between texts and images,
we specially design an image captioning task and a text-to-image generation task.
These two tasks are implemented by two decoders, which jointly regularize the embeddings from the encoders.
For the image captioning task,
the encoder takes the image embedding as input and generates the caption text,
while we do the opposite for the text-to-image generation task.
The overall procedure is summarized as follows.
First,
similar as in~\cite{zhang2021cross},
to enhance the generalization ability of the model,
we add random noise 
to the embeddings of texts and images:
\begin{align}
\mathbf{T_n} & = \text{MLP}(\mathbf{T_e} + \mathcal{N}_t(0,1)), \\
\mathbf{I_n} & = \text{MLP}(\mathbf{I_e} + \mathcal{N}_i(0, 1)), \label{decoder_noise}
\end{align}
where 
$\mathcal{N}_t(0,1)$ and $\mathcal{N}_i(0, 1)$ 
are Gaussian random noise 
for textual and visual modalities, respectively. 
Then we use two decoders to generate images $\hat{I}$ and texts $\hat{T}$, respectively: 
\begin{align}
\hat{I} & = \text{ImageDecoder}(\mathbf{T_n}), \\
\hat{T} & = \text{TextDecoder}(\mathbf{I_n}), \label{decoder}
\end{align}
where ImageDecoder can be a typical text-to-image generation model (e.g., PixelCNN~\cite{van2016conditional}), and TextDecoder can be a typical image captioning model (e.g., NIC~\cite{vinyals2015show}).  
Based on the generated images and texts, 
we compare them with the input ones and construct the text-to-image generation loss $\mathcal{L}_{ti}$ and image captioning loss $\mathcal{L}_{it}$, respectively.  
These two losses regularize the embeddings of texts and images generated by the encoders and can be considered as auxiliary tasks to the main prediction task that will be introduced later.


\subsection{Multimodal Exchanging}
After we derive embeddings $\mathbf{T_e}$ and $\mathbf{I_e}$
of textual and visual modalities, 
we next present the multimodal exchanging process. We start with the introduction of Transformer basics.

\paragraph{Transformer basics.}
Transformer~\cite{vaswani2017attention} is introduced for sequential data modeling with an encoder-decoder structure. 
In Transformer, 
the inputs are linearly mapped 
to three matrices: query $\mathbf{Q}$, key $\mathbf{K}$, and value $\mathbf{V}$. 
The multi-head self-attention is then computed by:
\begin{align}
\text{Attention}(\mathbf{Q}, \mathbf{K}, \mathbf{V}) = \text{Softmax}(\frac{\mathbf{Q}\mathbf{K}^T}{\sqrt{d_k}})\mathbf{V},
\label{attention}
\end{align}
where $\sqrt{d_k}$ is the scaling factor. 
Here,
$\text{Softmax}(\mathbf{Q}\mathbf{K}^T/\sqrt{d_k})$ generates 
an attention map matrix, which characterizes the attention scores between tokens. 
Generally,
Transformer adds
a special token \textit{cls} to the beginning of the input sequences, 
which is used to derive
the sentence-level embedding.
Therefore, 
the first row of the attention map represents the attention scores of \textit{cls} to all the tokens in the input. 
In this paper, 
we only use the Transformer encoder as in~\cite{kenton2019bert},
where 
each layer first computes the multi-head self-attention, followed by 
a feed-forward network (FFN) with residual connection~\cite{he2016deep} and layer normalization~\cite{ba2016layer}. 

\paragraph{\ct.}
Based on Transformer,
we propose 
\ct, 
which
uses two Transformer encoders with shared parameters to learn the embeddings of textual and visual modalities,
and exchanges information 
between multimodalities.
The overall pipeline of \ct\ is shown in component~\ding{175} of Figure~\ref{fig:main}.
We first add \emph{cls} to the beginning of 
the embeddings generated by the text encoder and the image encoder,
which are taken as the input of 
the \ct.
After that,
considering that the global contextual information of both input vectors should be first learned and then exchanged,
\ct\ sets its shallow layers as the regular Transformer encoder layer, 
followed by a number of exchanging layers. 
When multimodal fusion finishes,
the exchanging process stops.
To achieve this,
\ct\ introduces two hyper-parameters $\mu$ and $\eta$ to control the start layer and the end layer for information exchanging, respectively.
In each exchanging layer,
inspired by~\cite{caron2021emerging, liang2022not}, 
we select tokens in one modality with a proportion of the smallest attention scores to \emph{cls} and replace their embeddings vectors with the average embedding of all the tokens in the other modality.
Here,
\emph{cls} is taken as the reference because it generates the sentence-level embedding.




Specifically,
given the embeddings $\mathbf{T_e}$ and $\mathbf{I_e}$ generated from the text and image encoders,
we first add \textit{cls} to the beginning of them
and 
obtain the inputs 
$\mathbf{T}_{e}(0), \mathbf{I}_{e}(0) \in \mathbb{R}^{(n+1)\times d}$
for \ct: 
\begin{align}
\mathbf{T}_{e}(0) =
\texttt{Concat}\left (\mathbf{v}_{cls}(t_0), \mathbf{T}_{e}\right ), \\ \mathbf{I}_{e}(0) = \texttt{Concat} \left( \mathbf{v}_{cls}(i_0), \mathbf{I}_{e}\right), 
\end{align}
where 
$\mathbf{v}_{cls}(t_0)$ and $\mathbf{v}_{cls}(i_0)$ denote the initial embeddings 
of $cls$ for textual and visual modalities, respectively. 
Here, $n$ is the input text length and $d$ is the embedding dimensionality.
After $\mu$
regular Transformer encoder layers, 
we can obtain the updated embeddings $\mathbf{T}_{e}(\mu)$ and $\mathbf{I}_{e}(\mu)$. 
At layer $\mu+1$, 
we step into the exchanging layer,
which consists of three sub-modules.
The first sub-module 
calculates the
multi-head self-attention in Equation~\ref{attention},
which generates 
intermediate embeddings $\tilde{\mathbf{T}}_{e}(\mu+1)$ for texts and $\tilde{\mathbf{I}}_{e}(\mu+1)$ for images.
The second sub-module selects a $\theta$-proportion of tokens with the smallest attention scores to \emph{cls} for both modalities and performs information exchanging.
Suppose the token corresponding to the $k$-th row of $\tilde{\mathbf{T}}_{e}(\mu+1)$
is selected, 
whose embedding vector is then updated by:
\begin{align}
\begin{split}
\tilde{\mathbf{T}}_{e}(\mu+1)[k,:] & =  \frac{1}{n}\sum^{n}_{j=1}\tilde{\mathbf{I}}_{e}(\mu+1)[j,:] \\
& + \tilde{\mathbf{T}}_{e}(\mu+1)[k,:], 
\end{split}
\end{align}
where 
we employ residual connection~\cite{he2016deep}
to reduce the information loss caused by replacement.
Similarly, 
we can update the embedding vector in $\tilde{\mathbf{I}}_e(\mu+1)$ by:
\begin{align}
\begin{split}
\tilde{\mathbf{I}}_{e}(\mu+1)[k,:] & = \frac{1}{n}\sum_{j=1}^{n}\tilde{\mathbf{T}}_{e}(\mu+1)[j,:] \\
& + \tilde{\mathbf{I}}_{e}(\mu+1)[k,:].
\end{split}
\end{align}
After information exchange,
the third sub-module inputs the updated embedding matrices into FFN with layer normalization to obtain the output embeddings at layer $\mu+1$:
\begin{align}
\mathbf{T}_{e}(\mu+1) = \text{FFN}(\tilde{\mathbf{T}}_{e}(\mu+1)),\\ \mathbf{I}_{e}(\mu+1) = \text{FFN}(\tilde{\mathbf{I}}_{e}(\mu+1)).
\end{align}
The exchanging process continues until reaching the preset end layer $\eta$. 
Finally we concatenate the output embeddings of the two Transformers
and feed them into a fully connected network to derive the final fusion embedding 
matrix $\mathbf{F}_e$.

\subsection{Training Objective}
The overall optimization objective function of \muse\ is summarized as
\begin{align}
    \mathcal{L} = \mathcal{L}_{task} + \alpha \mathcal{L}_{it} + \beta \mathcal{L}_{ti},
\end{align}
where $\mathcal{L}_{task}$, $\mathcal{L}_{it}$, $\mathcal{L}_{ti}$ are the loss of the main task (MNER or MSA), the image captioning task and the text-to-image generation task, respectively. Here, 
$\alpha$ and $\beta$ are two hyper-parameters to balance the importance of the three terms. 

\section{Experiments}
\label{sec:exp}
In this section,
we evaluate the performance of \muse\ by comparing it with other multimodal fusion methods on two tasks: Multimodal Named Entity Recognition (MNER) and Multimodal Sentiment Analysis (MSA). 
Given an input pair of text $T$ and image $I$, 
MNER aims to extract a set of entities from $T$ and classify each entity into one of the pre-defined types;
the target of MSA is to classify each pair to one of the pre-defined sentiment types.


\paragraph{Common Implementation.}
Before diving into the details of experiments, we introduce the common implementation in two tasks. We use ResNet~\cite{he2016deep}, BERT~\cite{kenton2019bert}, PixelCNN++~\cite{salimans2017pixelcnn++}, NIC-Att~\cite{xu2015show} for ImageEncoder, TextEncoder, ImageDecoder, TextDecoder, respectively. The batch size is set to $40$, and the learning rate is selected via a small grid search [1e-4, 5e-5, 1e-5] on the validation set. In ImageEncoder, the input image is resized to [224, 224] and the encoded image size is set to [8, 8]. In TextEncoder, the max sequence length is set to $64$. In ImageDecoder, the input image is resized to [32, 32]. In \ct, we keep the same setting with Transformer encoder, except that the dropout rate is selected via a small grid search [0.1, 0.2, 0.3, 0.4] on the validation set. For loss selection, we use CrossEntropy for MNER prediction, MSA prediction and image captioning, and  use discretized mix logistic loss for text-to-image generation as in PixelCNN++. For initialization of \textit{cls}, we use Kaiming Initialization~\cite{he2015delving}. Our framework is implemented based on PyTorch~\cite{paszke2019pytorch}. We run all experiments on $6$ Tesla V100 GPUs. 

Due to the space limitation, we move the description of datasets to Appendix~\ref{datasets_appendix}.

\subsection{Multimodal Named Entity Recognition}


\paragraph{Implementation.} 
Following previous works~\cite{yu2020improving},
we feed the generated fusion embedding $\mathbf{F}_e$
into a CRF~\cite{lafferty2001conditional} layer
to make prediction for MNER. 
The learning rate and the dropout rate of CRF are set to $0.0001$ and $0.5$, respectively.  

\paragraph{Results.}
We compare our method with three types of competitors:
aggregation-based methods (GVATT~\cite{lu2018visual}, AdaCAN~\cite{zhang2018adaptive}, UMT~\cite{yu2020improving}), alignment-based methods (ITA~\cite{wang2021ita}), and task-specific methods (RIVA~\cite{sun2020riva}, RpBERT~\cite{sun2021rpbert}, UMGF~\cite{zhang2021multi}).
Details of these approaches 
can be found in Section~\ref{related_works_mner}.
We take precision, recall and F1 score as the evaluation measures.
For all the measures,
the larger value 
indicates the better model performance.
Table~\ref{tab:res_mner} summarizes the comparison results. 
From the table, we see that 
\muse\ achieves the best performance w.r.t. all the evaluation metrics on all the  datasets. 
In particular, \muse\ leads
the runner-up w.r.t. the F1 score 
by percents of $1.29$ and $1.22$ 
on Twitter15 and Twitter17, respectively. 
These results demonstrate the  effectiveness of our exchanging-based method for multimodal fusion. 
Further, we notice that
in MT-product, our method is the only one that can beat BERT-CRF, 
which also justifies the {generalizability} of \muse. 

Due to the space limitaion, we move the efficiency analysis on MNER datasets to Appendix~\ref{appe:eff_comp}.

\begin{table*}[h]
  \caption{MNER results (\%) w.r.t. precision (P), recall (R) and F1 scores. We highlight the best results in bold. * indicates that the improvements are statistically significant for $p< 0.01$ with paired t-test.}
  \label{tab:res_mner}
  \centering
  \begin{tabular}{c|ccc|ccc|ccc}
  \toprule
   \multirow{2}*{Methods} & \multicolumn{3}{c}{Twitter15}  &  \multicolumn{3}{c}{Twitter17}  &  \multicolumn{3}{c}{MT-Product}\\
    & P & R & F1 & P & R & F1 & P & R & F1 \\
    \midrule
    BERT-CRF & 69.22 & 74.59 & 71.81 & 83.32 & 83.57 & 83.44 & 45.13 & 42.94 & 44.01 \\
    \midrule
    GVATT~\cite{lu2018visual}  & 73.96 & 67.90 & 70.80 & 83.14 & 80.38 & 81.87 & - & - & -\\
    AdaCAN~\cite{zhang2018adaptive} & 72.75 & 68.75 & 70.69 & 84.16 & 80.24 & 82.15 & 40.15  & 41.66  & 40.89\\
    UMT~\cite{yu2020improving} & 71.67 & 75.23 & 73.41  & 85.28 & 85.34 & 85.31 & 40.86 & \textbf{43.69}  & 42.22\\
    RIVA~\cite{sun2020riva} & - & - & 73.80 & - & - & 87.40 & - & - & - \\
    RpBERT~\cite{sun2021rpbert} & - & - & 74.40 & - & - & 87.40 & - & - & - \\
    UMGF~\cite{zhang2021multi} & 74.49 & 75.21 & 74.85 & 86.54 & 84.50 & 85.51 & 41.67 & 43.55 & 42.58  \\
    ITA~\cite{wang2021ita} & - & - & 75.52 & - & - & 85.96 & - & - & -  \\
    \midrule
    \muse & \textbf{76.13} & \textbf{77.51} & \textbf{76.81} & \textbf{88.92} & \textbf{88.32}  & \textbf{88.62} & \textbf{49.17}  & 40.34   & \textbf{44.32}$^{*}$   \\
  \bottomrule
  \end{tabular}
\end{table*}

\begin{table*}[h]
  \caption{MSA results (\%) w.r.t. accuracy and F1 scores. 
  We highlight the best results in bold. * indicates that the improvements are statistically significant for $p< 0.01$ with paired t-test.}
  \label{tab:res_msa}
  \centering
  \begin{tabular}{c|cc|cc}
  \toprule
   \multirow{2}*{Methods} & \multicolumn{2}{c}{MVSA-Single}  &  \multicolumn{2}{c}{MVSA-Multiple}\\
    & Accuracy & F1 & Accuracy & F1 \\
    \midrule
   CNN-Multi~\cite{cai2015convolutional} & 61.20 & 58.37 & 66.39 & 64.19 \\
   DNN-LR~\cite{yu2016visual} & 61.42 & 61.03  & 67.86 & 66.33 \\
   HSAN~\cite{xu2017analyzing} & - & 66.90 & - & 67.76 \\
   MultiSN~\cite{xu2017multisentinet} & 69.84 & 69.63 & 68.86 & 68.11 \\
   CoMN~\cite{xu2018co} & 70.51 & 70.01  & 69.92 & 69.83 \\
   MVAN~\cite{yang2020image} & 72.98 & 72.98  & 72.36 & 72.30\\
   ITIN~\cite{zhu2022multimodal} & 75.19 & 74.97 & 73.52 & 73.49 \\
    \midrule
    \muse &  \textbf{75.80}$^{*}$ & \textbf{75.58}$^{*}$ & \textbf{74.10}$^{*}$  & \textbf{73.93}$^{*}$ \\
  \bottomrule
  \end{tabular}
\end{table*}

\subsection{Multimodal Sentiment Analysis}


\paragraph{Implementation.}
We apply a one-layer MLP 
on the fusion embedding of the \emph{cls} token
to classify the sentiments in MSA. The dropout rate of the MLP is set to $0.5$.

\paragraph{Results.}
We compare \muse\ with two types of approaches:
aggregation-based methods (CNN-Multi~\cite{cai2015convolutional}, DNN-LR~\cite{yu2016visual}, HSAN~\cite{xu2017analyzing}, 
MultiSN~\cite{xu2017multisentinet},
MVAN~\cite{yang2020image}), and alignment-based methods (ITIN~\cite{zhu2022multimodal}, CoMN~\cite{xu2018co}),
whose details are provided in Section~\ref{related_works_msa}.
We choose accuracy and F1 score as evaluation metrics.
For both metrics,
the larger the value, 
the better the model performance.
The results are given in Table~\ref{tab:res_msa}.
From the table,
we observe that 
\muse\ outperforms other competitors on both datasets. 
In particular,
compared with the runner-up model ITIN that is based on inter-modal alignment,
\muse\ utilizes both the intra-modal  processing and inter-modal exchanging-based fusion, which explains its superiority.


\subsection{Ablation Study}
We further conduct an ablation study to verify the importance of the main components in \muse. 
Specifically,
one variant takes only the textual input. 
We call this variant \textbf{\muse-only-$\mathbf{T}$}. 
Similarly,
we define another variant \textbf{\muse-only-$\mathbf{I}$} that inputs only the vision modality. 
These two variants help us recognize the importance of multimodal fusion for downstream tasks. 
To further show the significance of \ct, 
we remove it from \muse\ and directly employ the embeddings generated from TextEncoder and ImageEncoder.
We call this variant \textbf{\muse-w/o-CT}. 
Finally, 
To show the necessity of the image captioning task and the text-to-image generation task for embedding regularization, 
we propose three variants, 
in which 
\textbf{\muse-only-$\mathbf{\mathcal{L}_{task}}$} only optimizes the task loss, \textbf{\muse-w/o-$\mathbf{\mathcal{L}_{it}}$} removes the image captioning loss, and \textbf{\muse-w/o-$\mathbf{\mathbf{\mathcal{L}_{ti}}}$} removes the text-to-image generation loss. 

Table~\ref{tab:res_abl} summarizes the results w.r.t. the F1 score on both MNER and MSA tasks.
From the table, 
we see that: 
1) 
\muse\ clearly outperforms \muse-only-$T$ and \muse-only-$I$, which shows the importance of multimodal fusion 
for textual and visual representation in the MNER and MSA tasks. 
2) 
\muse\ achieves better performance than \muse-w/o-CT.
This is because \ct\ performs 
multimodal knowledge exchanging,
which boosts the quality of the learned embeddings for each modality. 
3) 
The outperformance of \muse\ over the remaining three variants shows 
the necessity of 
embedding regularization.
Since 
multimodal data could be projected into different spaces,
both
the image captioning task and the text-to-image generation task 
are used to pull embeddings of multimodalities into the same space,
based on which multimodal fusion can then be performed.
To conclude, all the components in \muse\ are important.

\begin{table*}[h]
  \caption{Ablation study results (\%) on \muse\ w.r.t. the F1 score. We highlight the best results in bold. }
  \label{tab:res_abl}
  \centering
  \begin{tabular}{c|ccccc}
  \toprule
   Methods & Twitter15 & Twitter17 & MT-Product & MVSA-Single  & MVSA-Multiple \\
   \midrule
   \muse-only-$T$ & 72.05  & 84.32  & 43.65  & 63.85  & 61.97  \\
   \muse-only-$I$ & - & -  & -  & 64.99  & 62.39 \\
   \muse-w/o-CT & 73.26  & 85.46 & 43.78  & 74.24   & 72.35 \\
   \muse-only-$\mathcal{L}_{task}$  & 74.39  & 87.63  & 44.20  & 74.66 & 72.95 \\
   \muse-w/o-$\mathcal{L}_{it}$ & 75.43  & 88.41  & 44.02  & 75.23 & 73.46 \\
   \muse-w/o-$\mathcal{L}_{ti}$ & 75.27   & 88.10  & 44.13  & 75.44  & 73.65  \\
    \midrule
    \muse &  \textbf{76.81} & \textbf{88.62} & \textbf{44.32} & \textbf{75.58}  & \textbf{73.93} \\
  \bottomrule
  \end{tabular}
\end{table*}

\begin{figure*}
     \centering
     \begin{subfigure}{0.32\textwidth}
        \centering
        \includegraphics[height=5cm]{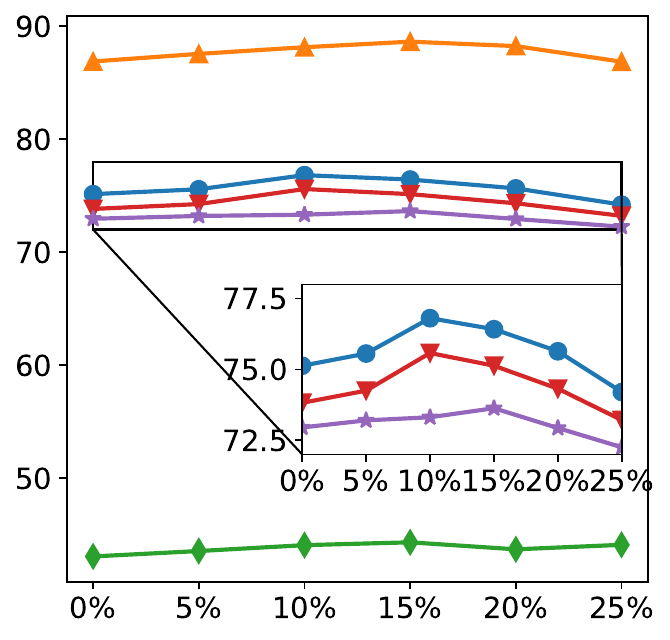}
        \caption{$\theta$}
        \label{fig:hyp_theta}
     \end{subfigure}
     \hfill
     \begin{subfigure}{0.32\textwidth}
        \centering
        \includegraphics[height=5cm]{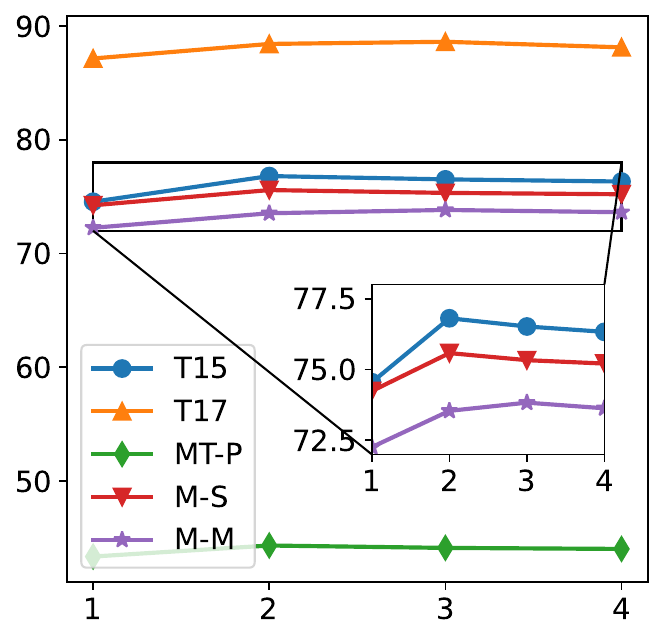}
        \caption{$\mu$}
         \label{fig:hyp_lambda}
     \end{subfigure}
     \hfill
     \begin{subfigure}{0.32\textwidth}
        \centering
        \includegraphics[height=5cm]{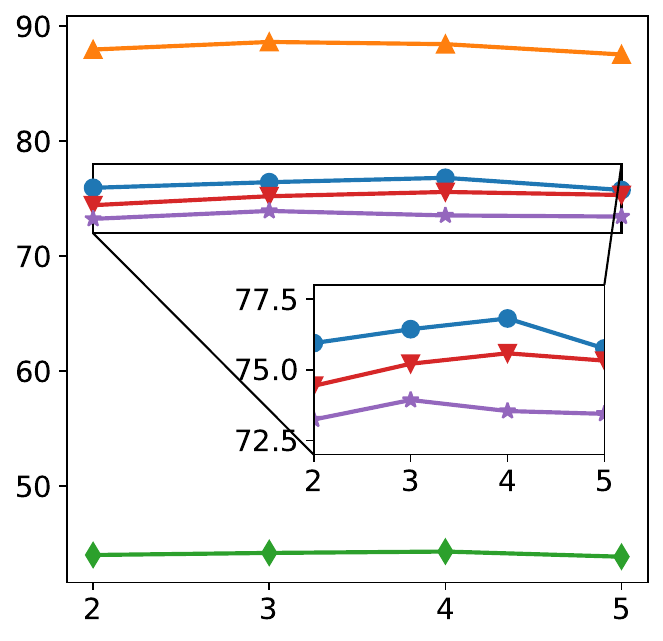}
        \caption{$\eta$}
         \label{fig:hyp_eta}
     \end{subfigure}
        \caption{
        Hyper-parameter sensitivity analysis on the exchange proportion $\theta$, the start layer $\mu$ and the end layer $\eta$ for information exchanging in \ct.
        We short
        Twitter15, Twitter17, MT-Product, MVSA-Single, and MVSA-Multiple 
        as T15, T17, MT-P, M-S, and M-M, respectively. 
        }
        \label{fig:hyp}
\end{figure*}


\subsection{Hyper-parameter Sensitivity Analysis}
We end this section with a sensitivity analysis of hyper-parameters in our model.
In particular,
we 
investigate three key hyper-parameters in \ct:
the exchange proportion $\theta$ of tokens, 
the start layer $\mu$ and the end layer $\eta$ for multimodal exchanging. 
We study one hyper-parameter with others fixed.
Figure~\ref{fig:hyp} shows the
experimental results w.r.t. the F1 score on all the datasets.
From the figure,
our observations can be summarized as follows.
 

\paragraph{The exchange proportion $\theta$.}
As shown in Figure~\ref{fig:hyp_theta},
the hyper-parameter $\theta$ plays an important 
role in \muse. 
We see that 
the model performance first increases and then drops,
with the increase of $\theta$.
When the $\theta$ is too small, the inter-modal fusion could be  inadequate,
which adversely affects the model performance. 
Further,
when $\theta$ is too large,
a large number of tokens will be replaced,
which might attenuate the raw intra-modal knowledge.
Our results empirically 
suggest that setting $\theta$ to be about $10\%$ to $20\%$
 could be a reasonable choice. In our experiments, the default setting of $\theta$ is $10\%$.

\paragraph{The start layer $\mu$.}
Since the default number of layers in the regular Transformer is 6, 
we vary the value of $\mu$ from $\{1, 2, 3, 4\}$.
For any $\mu$ value,
we fix
the end layer $\eta$ to $4$.
From Figure~\ref{fig:hyp_lambda},
we observe that the model performance improves quickly 
when $\mu$ increases from $1$ to $2$.
A possible explanation is that 
the randomly initialized embeddings for 
\textit{cls} might not capture the sentence-level semantics
and 
better contextual representations
can be learned with few layers.
This further demonstrates the 
necessity to use the regular Transformer layer without exchanging for contextual learning in
the shallow layers of \ct. 

\paragraph{The end layer $\eta$.}
We fix the start layer $\mu$ to 2 and show the results w.r.t. the end layer $\eta$ in Figure~\ref{fig:hyp_eta}.
Similar as in Figure~\ref{fig:hyp_lambda},
we also see that 
the model performance first rises and then degenerates, 
as $\eta$ increases.
Within exchanging layers,
\muse\ fuses knowledge extracted from one modality to the other,
which improves the quality of the learned representations and explains the performance increase.
However,
after the fusion completes,
more exchanging layers 
could degrade the original intra-modal knowledge, 
which could lead to performance degeneration.
We empirically set the default setting of $\eta$ to $4$ in the experiments.


\section{Conclusion}
In this paper, 
we proposed 
a novel exchanging-based model \muse\ for 
multimodal fusion.
\muse\ first uses two encoders
to project inputs of texts and images into separate low-dimensional spaces. 
Then
it regularizes the embeddings and pulls them into the same space with two generative decoders,
which capture the correlations between texts and images by two tasks: image captioning and text-to-image generation.
After that,
based on the regularized embeddings, 
we proposed \ct\ for multimodal fusion by exchanging token embeddings from different modalities.
We conducted extensive experiments to evaluate the performance of \muse\ on  MNER and MSA tasks.
Experimental results show that 
our method performs favorably against other SOTAs.

\newpage
\section{Limitation}
In this paper, we propose a new exchange-based multimodal fusion method that does better on the MNER and MSA tasks. Moreover, the hyperparameters of the exchange module are analyzed in detail. In the future, we will extend this method to more tasks.

\section{Ethical Considerations}
The proposed method has no obvious potential risks. All the scientific artifacts used/created are properly cited/licensed, and the usage is consistent with their intended use. Also, we open up our codes and hyper-parameters to facilitate future reproduction without repeated energy cost.

\bibliography{anthology,custom}
\bibliographystyle{acl_natbib}

\clearpage
\appendix

\section{Related Works}
\label{related_works_appendix}

\subsection{Multimodal Named Entity Recognition (MNER)}
\label{related_works_mner}

Existing works for 
MNER can be mainly categorized into three types: aggregation-based, alignment-based and task-specific. 
For example,
AttMNER~\cite{moon2018multimodal},  
GVATT~\cite{lu2018visual}
and AdaCAN~\cite{zhang2018adaptive} are pioneering works that employ 
the attention mechanism to aggregate text and image features.  
UMT~\cite{yu2020improving} further uses Transformer to enhance the aggregation performance.
The alignment-based method
ITA~\cite{wang2021ita} introduces three types of alignments between texts and images to improve the classification results. 
There are also some task-specific works, such as RIVA~\cite{sun2020riva} and RpBERT~\cite{sun2021rpbert}, which aim to solve the mismatch problem 
between texts and images. 
Further,
UMGF~\cite{zhang2021multi} proposes a unified multimodal graph fusion based on fine-grained graph modeling.

\subsection{Multimodal Sentiment Analysis (MSA)}
\label{related_works_msa}
MSA has recently attracted growing attention for its capability in learning cross-domain knowledge for sentiment analysis. 
Similar as in MNER,
there are also
aggregation-based methods for MSA. For example,
CNN-Multi~\cite{cai2015convolutional} 
first uses three CNNs to learn visual and textual features and then aggregates them for sentiment prediction. Further, DNN-LR~\cite{yu2016visual} introduces pre-trained CNNs to improve the performance. 
There also exist approaches~\cite{xu2017analyzing, xu2017multisentinet, yang2020image} that present hierarchical designs to enhance aggregation networks.
Another type of methods for MSA is based on aligning techniques.
For example,
CoMN~\cite{xu2018co} introduces mutual information to align the influences of texts and images; ITIN~\cite{zhu2022multimodal} utilizes cross-modal alignment to capture the region-word correspondence with the adaptive gating module, which has been shown to achieve superior 
results.

\section{Datasets}
\label{datasets_appendix}

\subsection{MNER}
We evaluate our method on two public datasets Twitter15~\cite{zhang2018adaptive} and Twitter17~\cite{lu2018visual},
and  
we keep the same Train/Validation/Test splits. 
The two datasets are comprised of multimodal user posts, 
which are 
collected from Twitter during 2014-2015 and 2016-2017, respectively. 
Each dataset contains four types of entities, namely person, location, organization and miscellaneous. We use the released version\footnote{Twitter15 and Twitter17 download url: \url{https://github.com/jefferyYu/UMT}.} from UMT~\cite{yu2020improving}. 
Additionally, 
we collected a product dataset MT-Product 
from a well-known 
E-commercial platform,
which
has only one type of entity, 
namely product.
In this dataset,
we only keep one sample for each product to ensure that products in the training and test sets do not overlap,
to further test the model generalizability.
We randomly split the dataset into the  Training/Validation/Test sets with the split ratio 
6/2/2. 
We will release this dataset later.
The statistics of the three datasets are shown in Table~\ref{tab:stat}.

\subsection{MSA}

We evaluate our method on two public datasets MVSA-Single and MVSA-Multiple\footnote{MVSA-Single and MVSA-Multiple download url: \url{https://mcrlab.net/research/mvsa-sentiment-analysis-on-multi-view-social-data/}.},
which are released in~\cite{niu2016sentiment}.
The two datasets are both collected from Twitter. For fair comparison, we preprocess the datasets following works~\cite{xu2017multisentinet, zhu2022multimodal}: 1) Remove the pairs whose image label and text label are inconsistent. 2) Split the datasets into the Training/Validation/Test sets randomly with the split ratio 8/1/1. Each dataset has three sentiment types, namely \emph{positive}, \emph{neutral}, and \emph{negative}. 
The statistics of the two datasets are shown in Table~\ref{tab:stat}.
\begin{table*}[h]
  \caption{The statistics of datasets.}
  \label{tab:stat}
  \centering
  \begin{tabular}{ccccc}
    \toprule
    Task     & Dataset     & Training & Validation & Test \\
    \midrule
     \multirow{3}*{Multimodal Named Entity Recognition} & Twitter15 & 4000 & 1000 & 3257\\
     & Twitter17 & 3373 & 723 & 723 \\
     & MT-Product & 2680 & 893 & 895 \\
    \midrule
     \multirow{2}*{Multimodal Sentiment Analysis} & MVSA-Single & 3600 & 440 & 471\\
     & MVSA-Multiple &  13600 & 1720 & 1707\\
    \bottomrule
  \end{tabular}
\end{table*}

\section{Efficiency Analysis}
\label{appe:eff_comp}
To evaluate the efficiency of \muse, we compare it to several representative cross-attention-based text-vision fusion methods from the number of parameters (\#Params), performance (Perf), training time in one epoch (TT), and evaluation time (ET). Table~\ref{tab:eff_comp} gives the detailed comparsion results. The results show that \muse\ significantly improves performance on both datasets but also comes with higher time costs. Further, we can observe that the time cost of \muse\ mainly comes from the exchanging module. We have rewritten the transformer layer of PyTorch but have not optimized it well. In the future, we will further optimize the code implementation. 

\begin{table*}[h]
  \caption{Efficiency comparison.}
  \label{tab:eff_comp}
  \centering
  \begin{tabular}{c|ccc|ccc}
  \toprule
   \multirow{2}*{Models(\#Params)} & \multicolumn{3}{c}{T15}  &  \multicolumn{3}{c}{T17}\\
    &Perf & TT & ET & Perf & TT & ET \\
    \midrule
    ViLBERT + bilstm-crf (210M) &    73.88  &     90s   &   12s    &  87.23   &     78s   &   8s     \\
UMT (208M)                        &   73.41    &    86s  &     10s  &  85.31   &   74s    &    7s   \\
MuSE w/o two tasks (204M) &    74.39   &  145s  &    28s   &  87.63   &   118s   &   27s  \\
MuSE (236M)                            &    76.13   &  180s   &    32s   &  88.62   &   161s  &  30s    \\
  \bottomrule
  \end{tabular}
\end{table*}

\end{document}